\documentclass[twoside]{article}
\usepackage{amsfonts,amssymb,amsbsy,textcomp,marvosym,picins,amsmath,caption,threeparttable,amsthm,subfigure,float,lastpage,lscape}
\usepackage{eurosym,mathrsfs,fancyhdr,CJK,multicol,graphics,indentfirst,color,bm,upgreek,booktabs,graphicx,multirow,warpcol}
\usepackage{epstopdf}
\usepackage{pifont}
\usepackage[normalem]{ulem}
\usepackage{booktabs}
\usepackage{makecell}
\usepackage{url}
\usepackage{latexsym}
\usepackage{amsmath}
\usepackage{graphicx}
\usepackage[subfigure]{graphfig}
\usepackage{multirow}
\usepackage{balance}
\usepackage{multirow, booktabs}
\usepackage{footnote}
\usepackage{enumitem, array}

\newtheorem{rmk}{Remark}
\newtheorem{definition}{Definition}

\def\footnoterule{\kern 1mm \hrule width 10cm \kern 2mm}

\allowdisplaybreaks
\catcode`@=11
\def\title#1{\vspace{3mm}\begin{flushleft}\vglue-.1cm\Large\bf\boldmath\protect\baselineskip=18pt plus.2pt minus.1pt #1
\end{flushleft}\vspace{1mm} }
\def\author#1{\begin{flushleft}\normalsize #1\end{flushleft}\vspace*{-4pt} \vspace{3mm}}
\def\address#1#2{\begin{flushleft}\vglue-.35cm${}^{#1}$\small\it #2\vglue-.35cm\end{flushleft}\vspace{-2mm}\par}
\def\jz#1#2{{$^{\footnotesize\textcircled{\tiny #1}}$\let\thefootnote\relax\footnotetext{\!\!$^{\footnotesize\textcircled{\tiny #1}}$#2}}}
\catcode`@=11
\def\section{\@startsection{section}{1}{\z@}%
 {-3ex \@plus -.3ex \@minus -.2ex}%
 {2.2ex \@plus.2ex}%
{\normalfont\normalsize\protect\baselineskip=14.5pt plus.2pt minus.2pt\bfseries}}
\def\subsection{\@startsection{subsection}{2}{\z@}%
 {-3ex\@plus -.2ex \@minus -.2ex}%
 {2ex \@plus.2ex}%
{\normalfont\normalsize\protect\baselineskip=12.5pt plus.2pt minus.2pt\bfseries}}
\def\subsubsection{\@startsection{subsubsection}{3}{\z@}%
 {-2.2ex\@plus -.21ex \@minus -.2ex}%
 {1.4ex \@plus.2ex}
{\normalfont\normalsize\protect\baselineskip=12pt plus.2pt minus.2pt\sl}}

\begin{document}
\begin{CJK*}{GBK}{song}
\thispagestyle{empty}
\vspace*{-13mm}
\vspace*{2mm}

\title{A Communication Theory Perspective on Prompting Engineering Methods for Large Language Models}

\author{Yuanfeng Song$^{1}$, Yuanqin He$^{1}$, Xuefang Zhao$^{1}$, Hanlin Gu$^{1}$, Di Jiang$^{1}$, Haijun Yang$^{1}$, Lixin Fan$^{1}$, and Qiang Yang$^{1}$}

\address{1}{AI Group, WeBank Co., Ltd, China}

\vspace{2mm}

\noindent E-mail: \{yfsong, yuanqinhe, summerzhao, allengu, dijiang, navyyang, lixinfan, qiangyang\}@webank.com;\\[-1mm]

\noindent {\small\bf Abstract} \quad  {\small {The springing up of \emph{Large Language Models} (LLMs) has shifted the community from single-task-orientated natural language processing (NLP) research to a holistic end-to-end multi-task learning paradigm. Along this line of research endeavors in the area, LLM-based \textit{prompting methods} have attracted much attention, partially due to the technological advantages brought by prompt engineering (PE) as well as the underlying NLP principles disclosed by various prompting methods. 
Traditional supervised learning usually requires training a model based on labeled data and then making predictions. 
In contrast, PE methods directly use the powerful capabilities of existing LLMs (i.e., GPT-3 and GPT-4) via composing appropriate prompts, especially under few-shot or zero-shot scenarios. Facing the abundance of studies related to the prompting and the ever-evolving nature of this field, this article aims to (i) illustrate a novel perspective to review existing PE methods, within the well-established communication theory framework; (ii) facilitate a better/deeper understanding of developing trends of existing PE methods used in four typical tasks; (iii) shed light on promising research directions for future PE methods.}}

\vspace*{3mm}


\vspace*{4mm}

\end{CJK*}
\baselineskip=18pt plus.2pt minus.2pt
\parskip=0pt plus.2pt minus0.2pt
\begin{multicols}{2}

\section{Introduction}

Large Language Models (LLMs) (e.g., GPT-3 \cite{brown2020language}, GPT-4 \cite{GPT4}, LLaMa \cite{touvron2023llama}) make it possible for machines to understand users' attention accurately, thus revolutionizing the human-computer interaction (HCI) paradigm. Compared to traditional machine systems like databases and search engines, LLMs demonstrate impressive capability in understanding, generating, and processing natural language, facilitating a series of services ranging from personal assistants \cite{cheng2023potential}, healthcare \cite{cascella2023evaluating} to e-commercial tools \cite{george2023review} via a {unified} natural language interface between users and machine. 
 
The research paradigm around LLM has shifted from single-task-orientated natural language processing (NLP) research to a holistic end-to-end multi-task learning approach. Along this line of research endeavors, LLM-based \textit{prompting engineering} (PE) methods \cite{liu2023pre,brown2020language} have attracted much attention, partially because they are the key techniques in making full use of the superior capabilities of LLMs via constructing appropriate prompts. 
PE refers to the process of crafting effective instructions to guide the behavior of LLMs, and it greatly helps in bridging the gap between the pre-training tasks used to construct the LLM with the down-streaming tasks queried by the end users. Through careful prompt designing, users can steer LLM's output in the desired direction, shaping its style, tone, and content to align with their goals. 

\begin{figure*}[t!]
   \centering
   \includegraphics[width=0.65\textwidth]{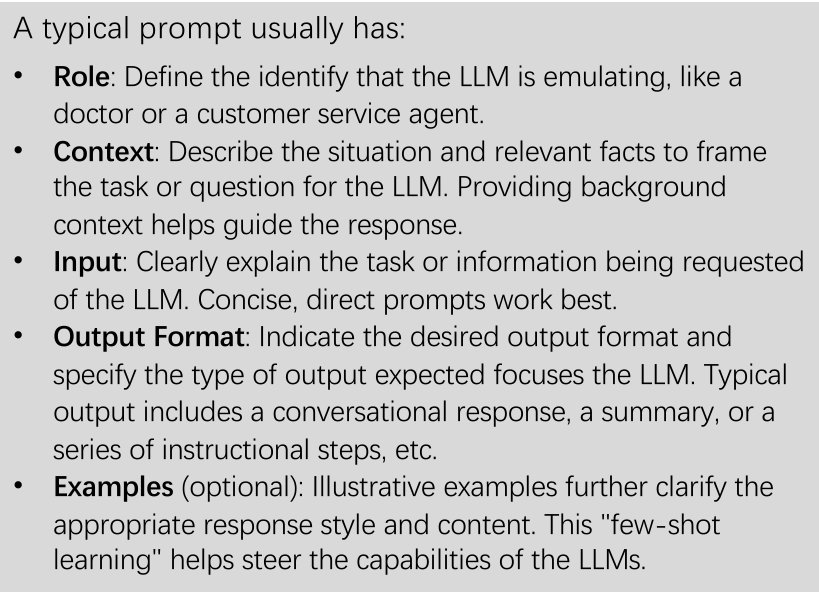}
   \caption{Components of A Good Prompt.}
   \label{fig:prompt}
\end{figure*}

To this end, numerous prompt engineering (PE) methods have been explored with the notable progress of LLM advancement and technologies \cite{liu2023pre,radford2019language,petroni2019language,schick2020exploiting,jiang2020can,shin2020autoprompt,li2021prefix,haviv2021bertese,liu2021gpt,zhong2021factual,gao2020making,zhang2021differentiable,han2022ptr,lester2021power,gu2021ppt,deng2022rlprompt,hou2022metaprompting,wang2023multitask}. A common theme of PE development lies in continuously improving accuracy and responsiveness of designed prompts, which often include components like Role, Context, Input, Output Format, and Examples (see Fig.~\ref{fig:prompt}).   Specifically, prompt template and answering engineering has evolved from solely utilizing discrete prompts to continuous prompts, and even to exploring hybrid prompts that combine continuous and discrete elements, which provides a larger optimization space to achieve better performance.
With the emergent capability of LLM, LLMs are leveraged to plan and use external tools via its in-context learning capability, which significantly enhanced its ability in specialized domains and broadened its application fields.

\begin{figure*}[t!]
   \centering
   \includegraphics[width=1.0\textwidth]{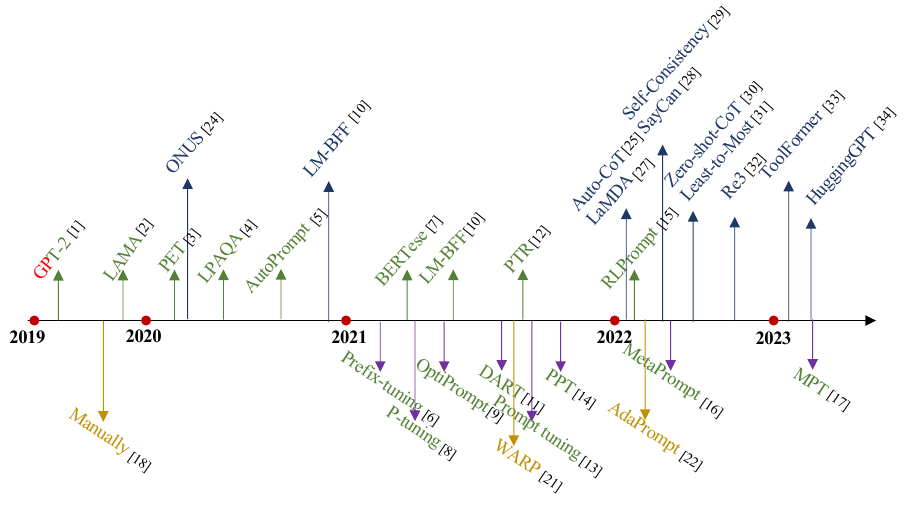}
   \caption{Chronological overview of representative studies in prompting methods from four aspects: prompt template engineering \cite{radford2019language,petroni2019language,schick2020exploiting,jiang2020can,shin2020autoprompt,li2021prefix,haviv2021bertese,liu2021gpt,zhong2021factual,gao2020making,zhang2021differentiable,han2022ptr,lester2021power,gu2021ppt,deng2022rlprompt,hou2022metaprompting,wang2023multitask}, prompt answering engineering \cite{yin2019benchmarking,petroni2019language,schick2020automatically,jiang2020can,schick2020exploiting,shin2020autoprompt,cui2021template,gao2020making,hambardzumyan2021warp,chen2022adaprompt, kim2022accurate}, and multi-turn prompting and multi-prompt learning \cite{schick2020exploiting,perez2020unsupervised,gao2020making,shum2023automatic,paranjape2023art,thoppilan2022lamda,ahn2022cana,wang2022self,kojima2023large,zhou2022leasttomost,yang2022re3,schick2023toolformer,shen2023hugginggpt}.}
   \label{fig:timeline}
\end{figure*}

Following these studies, one can summarize representative PE methods in a chronological overview as illustrated in Fig.~\ref{fig:timeline}. These methods can be categorized as three groups that respectively correspond to three prompting tasks proposed to improve the qualities of LLMs' outputs, namely \emph{prompt template engineering}, \emph{prompt answer engineering}, and \emph{multi-turn prompting and multi-prompt learning}. An example of the input and output for the above-mentioned tasks can be found in Table~\ref{tb:example}. 

\begin{itemize}
\item First, \textit{prompt template engineering} methods aim to carefully design a piece of ``text'' that guides the language models to produce the desired outputs. For example, in Table~\ref{tb:example}, to finish a classical sentiment detection for a input \emph{A=``Great places to eat near my location!''}, the prompt template engineering designs a template ``\emph{[A] Overall, it was a $[Z]$ restaurant}'' to enforce the LLM to fill the desired comments in the blank i.e. $[Z]$. Essentially this type of template engineering method induces LLM to focus on word embeddings that are relevant to the questions. A common designing principle of existing prompt template engineering methods is to better align information between users and LLMs. Such a trend is manifested by the evolution from using discrete prompts (e.g., a piece of human-readable text) \cite{jiang2020can, petroni2019language} to continuous ones (e.g., a continuous task-specific vector) \cite{li2021prefix, lester2021power}.

\item Second, \textit{prompt answer engineering} \cite{liu2023pre} refers to the process of searching for an answer space and a map to the original output, which enhances users' understanding of the information encapsulated within the LLM. For the same example in Table~\ref{tb:example}, the prompt answer engineering aims to find a mapping from the result \emph{``good''} obtained from the LLM to the desired answer \emph{``positive''}. The field of prompt answer engineering is currently witnessing a notable development trend characterized by the pursuit of models that excel in decoding model information from simple mapping to complex mapping to enhance human comprehension. 

\item Third, \textit{multi-prompting methods} mainly applied ensemble techniques \cite{schick2020exploiting} to mitigate the sensitivity of LLM to different formulations and to obtain a more stable output. In Table~\ref{tb:example}, the multi-prompting methods combine three different templates (i.e., \emph{1. ``It was a [Z]''; 2. ``Just [Z]''; 3. ``All in all, it was [Z]'';}) and their inference results (i.e., \emph{1. ``good'' 2. ``great!'' 3. ``okay''}) to obtain the final desired one (i.e., \emph{``positive''}). 
Later, as LLMs become more capable, multi-turn prompt methods attract more attention that aims to provide more context to LLM by leveraging information either from LLM itself or external tools \cite{kojima2023large, paranjape2023art}. 
In the field of multi-prompting methods, researchers are endeavoring to develop adaptive strategies that enhance LLM's ability to task planning and the utilization of tools. 
\end{itemize}

In this article, we summarize the prompting methods from a \textit{communication theory} perspective with which the ultimate goal of PE is to \emph{reduce the information misunderstanding between the users and the LLMs}. 
Therefore, as delineated in Section 2, the communication theory perspective provides a coherent explanation of different PE methods in terms of their objectives and underlying principles.   
Moreover, this novel perspective also offers and presents insights into scenarios where existing prompting methods come short.

\begin{table*}[th!]
\centering
\caption{Running Examples for PE Methods}
\begin{tabular}{>{\centering\arraybackslash}m{3cm}|m{6cm}|m{6cm}}
\hline
Stage & Input & Output  \\
\hline
\hline
Prompt Template Engineering & \emph{Great places to eat near my location!} & \emph{Great places to eat near my location!} Overall, it was a $[Z]$ restaurant. \\
\hline
Large Language Model & \emph{Great places to eat near my location!} Overall, it was a $[Z]$ restaurant. & \emph{Great places to eat near my location!} Overall, it was a \emph{good} restaurant. \\
\hline 
Prompt Answering Engineering & good & positive \\
\hline
Multi-Prompt & 1. It was a [Z]; 2. Just [Z]; 3. All in all, it was [Z]; & 1. good 2. great! 3. okay  \\
\hline
\end{tabular}
\label{tb:example}
\end{table*}

The remainder of the article is structured as follows: Section~\ref{sec:over} details the overview of the prompting methods from the communication theory perspective. 
Sections~~\ref{sec:template}, ~\ref{sec:answer}, and ~\ref{sec:multi} review and summarize the recent progresses, respectively, from four PE tasks namely prompt template engineering, answer engineering, and multi-turn prompting methods. 
Section~\ref{sec:dis} discusses other related surveys and potential research directions.
Finally, we conclude this article in Section~\ref{sec:con} by summarizing significant findings and discussing potential research directions.

\section{A Communication Theory Perspective of Prompting Methods}
\label{sec:over}

The study of modern communication theory, which dates back to the 1940s and the following decades, gave rise to a variety of communication models including both linear transmission models and non-linear models such as interaction, transaction, and convergence models \cite{narula2006handbook,chandler2011dictionary,cobley2013theories}. A common theme of these early studies is to analyze how individuals utilize verbal and non-verbal interactions to develop meaning in diverse circumstances. Conceptually, the communication process is often modeled as a chain of information processing steps involving encoding, transmitting, and decoding of messages, between a sender and a receiver.  

\begin{figure*}[t!]
  \centering
    \subfigure[{The classical \textit{Interaction Model of Communication}.}]{
   \centering
     \includegraphics[width=0.45\textwidth]{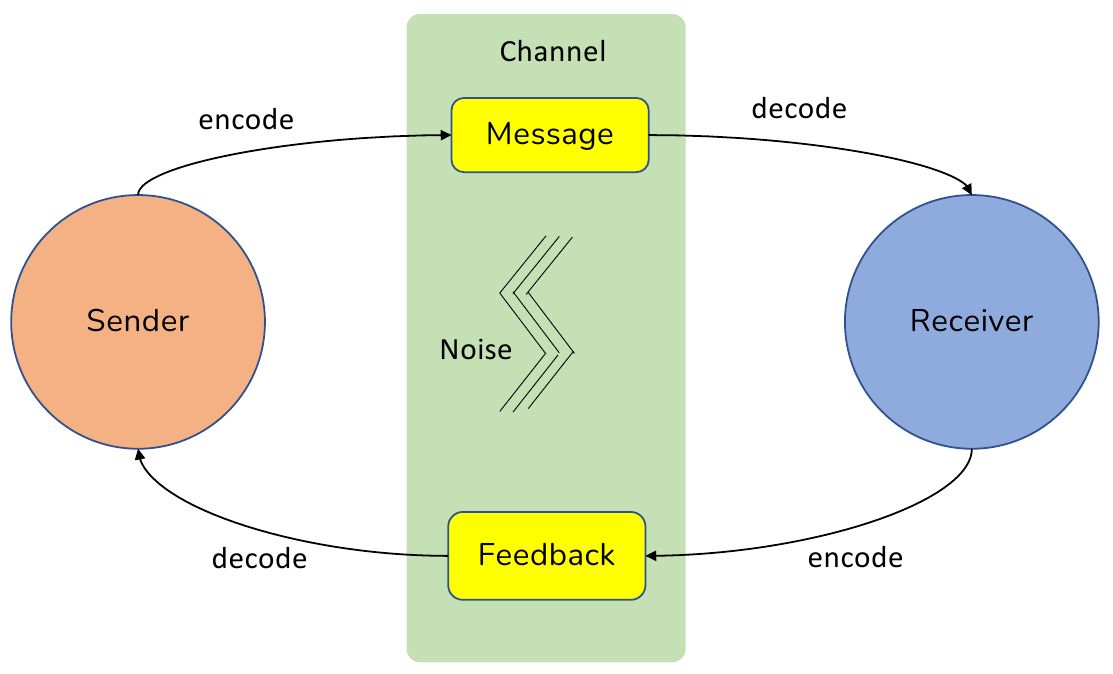}
     \label{fig:f1}
  }
   \subfigure[{Different Aspects of Existing Prompting Methods.}]{
   \centering
     \includegraphics[width=0.45\textwidth]{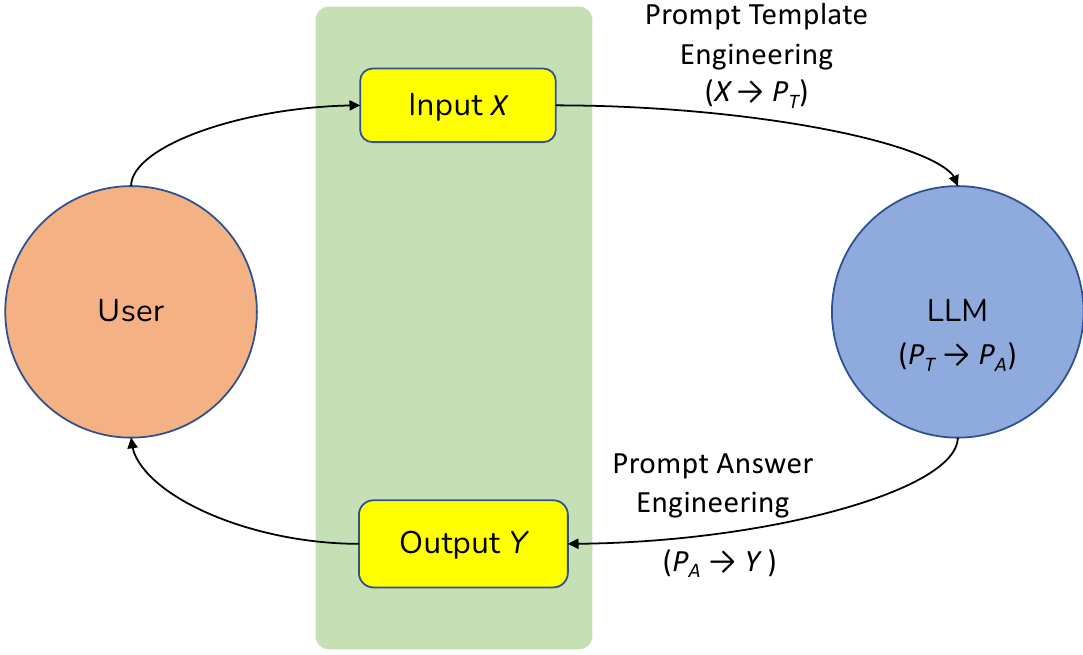}
     \label{fig:f2}
   }
  \caption{Prompting methods from the communication theory perspective.}
\label{fig:parameters}
\end{figure*}

To give a better illustration, Fig.~\ref{fig:f1} depicts the classical \textit{Model of Communication} in communication theory, which includes a sender encoding a message and transmitting it to the receiver over a channel. Then, the receiver decodes the message and delivers some type of response. During the transmission process, the message may be distorted due to noise, leading to the necessity of \textit{multi-turn} interaction.

The original communication theory is widely utilized to examine factors including social \cite{latane1996dynamic}, cultural \cite{orbe1998standpoint}, and psychological \cite{segrin1994negative} that influence human communication. 
The overall goal of communication theory is to reveal and clarify the common human experience of interacting with others through information exchange. 

Among early studies of various communication models, we are particularly inspired by two influential works, namely, Shannon-Weaver's mathematical model of communication \cite{shannon1948mathematical} and Schramm's communication model \cite{schram1954process}.  Shannon-Weaver's pioneering work, first published in 1948, provides a strong mathematical foundation to analyze information flow between an active sender and a passive receiver. It is however over-simplistic in the sense that it does not take into account of complexities involved in interactive communication between active senders and receivers, who may respond by sending their message as a form of feedback.   The interaction models of communication were first studied by Scharmm and published in his 1954 book \cite{schram1954process}, which pictorially illustrated the feedback loop as depicted in Fig.~\ref{fig:f1}.  Nevertheless, Scharmm's model falls short of rigorous theoretical and mathematical formulation to accommodate quantitative analysis e.g. information gain or mutual information between senders and receivers. 

Various prompting engineering methods for LLM, in our view, can be understood from Scharmm's model point of view (see Fig.~\ref{fig:f2}).  In the same vein of Shannon-Weaver's analysis, we, therefore, delineate a mathematical formulation of Prompting Engineering Systems for interactive user-LLM communication as follows: 

\begin{definition}
A Prompt Engineering System (PES) consists of a processing chain 
\begin{equation}
    X \stackrel{g_{\omega_T}}\longrightarrow P_T \stackrel{f_\theta}\longrightarrow P_A \stackrel{h_{\omega_A}}\longrightarrow Y,
\end{equation} 
where $g_{\omega_T}$ represents the mapping from the input $X$ to the prompt $P_T$, $f_\theta$ denotes the mapping from the prompt $P_T$ to the answer $P_A$ and $h_{\omega_A}$ denotes the mapping from the answer $P_A$ to the output $Y$ (see Fig.~\ref{fig:f2} for an illustration).
\end{definition}

\begin{definition}[Goal of PES] \label{def:cap-pes}
PES aims to maximize the mutual information between the inputs $X$ and outputs $Y$, i.e., 
\begin{equation}\label{eq:max}
    \max_{\omega_T, \omega_A}I(X,Y) = \max_{\omega_T, \omega_A}I(X,h_{\omega_A} \circ f_\theta \circ g_{\omega_T}(X))
\end{equation} 
where $f\circ g(x)=f(g(x))$.
\end{definition}

It's worth noting that prompt engineering is consistently divided into two procedures: Prompt Template Engineering and Prompt Answer Engineering. Each procedure has specific goals similar to Eq. \eqref{eq:max} that align with its intended purpose.

While the capacity in Def. \ref{def:cap-pes} is well-known in information theory \cite{cover1999elements}, how to reach the maximum of Eq. \eqref{eq:max} for large language models illustrated in Fig.~\ref{fig:f2} remains an unexplored research direction. There exists a large variety of prompting engineering methods, which, in our view, essentially aim to reduce information misunderstanding between users and LLMs. In other words, they aim to reach the capacity of PES as defined. This connection between PES and the communication models has never been explicitly stated before. 

Moreover, the existing work can be divided into three categories: prompt template engineering ($X \stackrel{g_{\omega_T}}\longrightarrow P_T$), prompt answer engineering ($P_A \stackrel{h_{\omega_A}}\longrightarrow Y$), and multi-prompt and multi-tune prompting as shown in Fig.~\ref{fig:f2}.
Specifically, the prompt template engineering aims to \textit{reduce the encoding error/ look for the prompt that is easily understood by the machine}, while the prompt answering engineering aims to \textit{reduce the decoding error/ look for the prompt that easily understood by the human}. The development of LLMs aims to \textit{enhance the capability of the receiver} that could better handle users' information needs, and most importantly, the multi-turn prompting and multi-prompt engineering aim to \textit{constantly reduce the information misunderstanding via multi-turn interactions}.

\begin{itemize}
    \item 
    \textbf{Prompt template engineering} aims to optimize
    \begin{equation}\label{eq:pt}
        \max_{\omega_T}I(X, P_A) =\max_{\omega_T}I(X,  f_\theta \circ g_{\omega_T}(X)) ,
    \end{equation}
    which looks for an additional piece of text, namely a prompt, to steer the LLMs to produce the desired outputs for downstream tasks. From the communication theory perspective, it acts as an ``encoder'' to bridge the gap between the users and the LLMs by encoding the messages in a way that the model can understand and then elicit knowledge from LLMs (see details in Section~\ref{sec:template}). 
    In the encoding process, the challenge lies in the accurate understanding of the user's intention by LLM with limited instruction following capability. Template engineering aims to reduce this mismatch by translating the user's request to a format that could be better understood by LLM. 
    \item \textbf{Prompt answer engineering}  aims to optimize
    \begin{equation}\label{eq:pa}
        \max_{\omega_A}I(P_T, Y) = \max_{\omega_A}I(P_T, h_{\omega_A}\circ f_\theta (P_T)),
    \end{equation}
    \noindent which focuses on developing appropriate inputs for prompting methods, has two goals: 1) search for a prompt answer $P_A$; 2) look for a map to the target output $Y$ that will result in an accurate predictive model. 
    In the decoding process, LLM-generated output often carries redundant information in addition to the expected answer due to its unlimited output space. Answer engineering aims to confine the output space and extract the target answer. The field of prompt answer engineering is currently witnessing a notable development trend characterized by the pursuit of effective answer engineering such that ultimate outputs (i.e. $Y$) are well aligned with that of end users' expectations (see details in Section~\ref{sec:answer})

    \item To further reduce the information misunderstanding, the user could conduct multi-interaction according to Eq. \eqref{eq:pt} and Eq. \eqref{eq:pa}, called \textbf{multi-prompt/multi-turn PE}.
    \textit{Multi-prompting methods} aims to optimize
    \begin{equation}\label{eq:mpt}
       \max_{\omega_{T_i}}\sum_{i=1}^M I(X, f_\theta \circ g_{\omega_{T_i}}(X)) ,
    \end{equation}
    which mainly applied ensemble techniques \cite{schick2020exploiting} to mitigate the sensitivity of LLM to different formulations and to obtain a more stable output. Later, as LLMs become more capable, multi-turn prompt methods
    focus to provide more context to LLM by leveraging multiple communication procedures between the machine and person  \cite{kojima2023large, paranjape2023art}.
     In the field of multi-prompting methods, researchers are endeavoring to develop adaptive strategies that enhance LLM's ability to task planning and the utilization of tools. The adaptive and iterative nature of multi-prompting methods is by the communication theory (see Section~\ref{sec:multi} for an elaborated explanation).
\end{itemize}

\section{Prompt Template Engineering}
\label{sec:template}

\begin{figure*}[t!]
   \centering
   \includegraphics[width=1.0\textwidth]{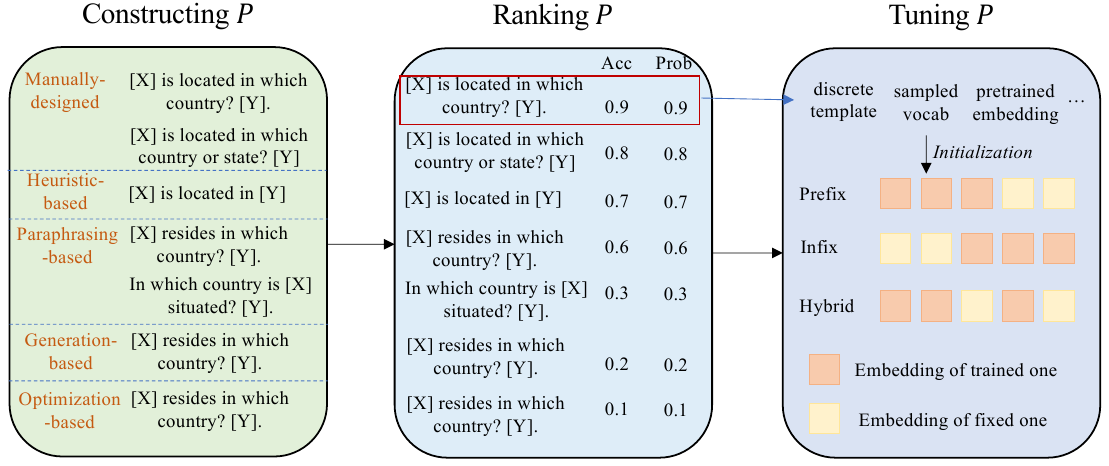}
   \caption{Overview of the PE Methods}
   \label{fig:pte}
\end{figure*}

Given the information chain $X \to P_T \to P_A$, the answer $P_A$ is determined by the prompt-processed $P_T$ and model $M$ with pre-trained weights $\theta$. Suppose that $\bar{P_A}$ is the targeted prediction, the key problem of prompt template engineering is to find a good prompt that maximizes the probability $p(\bar{P_A}|M, P_T, \theta)$ on diverse downstream tasks with limited data.
To obtain the optimal prompt, current works \cite{radford2019language,petroni2019language,schick2020exploiting,jiang2020can,shin2020autoprompt,li2021prefix,haviv2021bertese,liu2021gpt,zhong2021factual,gao2020making,zhang2021differentiable,han2022ptr,lester2021power,gu2021ppt,deng2022rlprompt,hou2022metaprompting,wang2023multitask} can be formulated into three categories: constructing $P_T$, ranking $P_T$ and tuning $P_T$, as shown in Fig.~\ref{fig:pte}.

\subsection{Constructing $P_T$}
The basic motivation of constructing $P_T$ is to transform the specific task to make it align with the pre-training objective (i.e., next-word prediction, masked LM) of the LM. As shown in Table~\ref{tb:construction}, existing prompt constructing methods \cite{petroni2019language,schick2020exploiting,schick2020exploiting,schick2020s,petroni2019language,radford2019language,liu2021gpt,raffel2020exploring,zhou2023revisiting,zhou2023revisiting,jiang2020can} could be categorized into five different approaches, which are discussed in detail as follows. 

\subsubsection{Manually-designed}
Initially, the prompt templates are manually designed in natural language based on the user's experience, and they have been validated to be able to improve the performances of downstream tasks, especially in a zero-shot setting \cite{brown2020language,radford2019language}. 
The most frequent style is to reformulate the original task as a 'fill-in-blank' cloze one \cite{petroni2019language,schick2020exploiting}, and the answer is obtained by predicting the words in the given [mask] place. 
For example, as illustrated in Fig.~\ref{fig:pte} and Table~\ref{tb:construction}, Petroni \textit{et al.} \cite{petroni2019language} manually designed prompts to re-structure the relational knowledge, while studies like \cite{schick2020exploiting, schick2020s} are dedicated to solving the text classification and language understanding tasks by several self-defining prompt patterns and propose a new training procedure named PET.  
Another line of work involves developing prefix prompts for generation tasks, which provide instructions and steer the LLMs to finish the sentence. 
For example, a summarization task can be handled by adding `TL;DR:' \cite{radford2019language}, and a translation task can be conducted into '\emph{Eng}. Translate to Spanish: \emph{Span}' \cite{raffel2020exploring}.
Even though manually designed prompts show some effectiveness \cite{zhou2023revisiting}, they are also criticized for being time-consuming and unstable \cite{liu2021gpt}. 
A subtle difference in the designed prompts may result in a substantial performance decrease. 
As such, how to explore the prompt space and construct prompts more thoroughly and more effectively becomes an important and challenging issue.

\begin{table*}[t!]
\caption{Summary of the prompt construction methods}
\centering
\label{PFT-table}
\begin{tabular}{c|c|c|c|c|c|c}
\hline
Method   & Automated & Gradient-Free  & Few-shot & Zero-shot & Stability & Interpret-ability \\ 
\hline
\hline
Manually-design \cite{radford2019language,petroni2019language,schick2020exploiting} &  
\ding{55} & \ding{51} & \ding{51}  &  \ding{51}  &  \ding{55} &  \ding{51} \\
Heuristic-based \cite{jiang2020can,logan2021cutting,han2022ptr} &  
\ding{51} & \ding{51} & \ding{51}  &  \ding{51}  &  \ding{51} &  \ding{51} \\
Paraphrasing-based \cite{jiang2020can,yuan2021bartscore,haviv2021bertese} &  
\ding{51} & \ding{51} & \ding{51}  &  \ding{51}  &  \ding{55} &  \ding{51} \\
Generation-based \cite{gao2020making,ben2021pada} &  
\ding{51} & \ding{51} & \ding{51}  &  \ding{51}  &  \ding{51} &  \ding{51} \\
Optimization-based \cite{shin2020autoprompt,deng2022rlprompt,zhou2022large} &  
\ding{51} & \ding{55} & \ding{51}  &  \ding{55}  &  \ding{51} &  \ding{55} \\
\hline
\end{tabular}
\label{tb:construction}
\end{table*}

\subsubsection{Heuristic-based}
The heuristic-based methods focus on finding acceptable prompts by some intuitive strategies. 
For example, to construct more flexible and diverse prompts for different examples (rather than the fixed ones), Jiang \textit{et al.} \cite{jiang2020can} propose to use the most frequent middle words and the phrase spanning in the shortest dependency path that appeared in the training data as a prompt. This method shows a large performance gain compared to the manually-designed prompts. Han \textit{et al.} \cite{han2022ptr} tries to form task-specific prompts by combining simple human-picked sub-prompts according to some logic rules. 
Different from the above methods, Logan \textit{et al.} \cite{logan2021cutting} uses an extremely simple uniform rule by \emph{null} prompts, which only concatenates the inputs and the [mask] token, and it's able to gain a comparable accuracy with manually-defined prompts.

\subsubsection{Paraphrasing-based}
The paraphrasing-based methods are widely used in data augmentation, aiming at generating augmented data that is semantically related to the original text, and this could be achieved in various ways using machine translation, model-based generation, and rule-based generation \cite{li2022data}. 
The paraphrasing-based methods could naturally be used to construct prompt candidates based on the original text, and we could further select the best one or integrate them to provide better performance. 
Representative studies includes \cite{jiang2020can, yuan2021bartscore, haviv2021bertese}. 
Specifically, Jiang \textit{et al.} \cite{jiang2020can} uses back-translation to enhance the lexical diversity while keeping the semantic meaning. 
Yuan \textit{et al.} \cite{yuan2021bartscore} manually creates some seeds and finds their synonyms to narrow down the search space. 
Haviv \textit{et al.} \cite{haviv2021bertese} uses a BERT-based model to act as a rewriter to obtain prompts that LLMs can understand better.

\subsubsection{Generation-based}
The generation-based methods treat prompt searching as a generative task that can be carried out by some LMs. For example, Gao \textit{et al.} \cite{gao2020making} first make use of the generative ability of T5 \cite{raffel2020exploring} to fill in the placeholders as prompts, and then the prompts could be further improved by encoding domain-specific information \cite{ben2021pada}.

\subsubsection{Optimization-based}
To alleviate the weakness of insufficient exploration space faced by existing methods, the optimized-based methods try to generate prompts guided by some optimization signals. For example, Shin \textit{et al.} \cite{shin2020autoprompt} employs gradients as the signals, and then searches for discrete trigger words as prompts to enrich the candidate space. Deng \textit{et al.} \cite{deng2022rlprompt} generates the prompt using a reinforced-learning approach that is directed with the reward function.

\subsection{Ranking $P_T$}

After obtaining multiple prompt candidates with the above-mentioned methods, the next step is to rank them to select the most effective one. 
Existing studies solve this problem by finding prompts that are close to the training samples to reduce the information mismatch between the pre-training and inference phases.

\subsubsection{Execution Accuracy}
Since the objective of the designed prompts is to fulfill the downstream tasks, it's intuitive and straightforward to evaluate the performance by execution accuracy over the specific tasks \cite{zhou2022large,gao2020making,jiang2020can}.

\subsubsection{Log Probability}
The log probability criterion prefers the prompt that delivers the correct output with higher probability, rather than being forced to give the exact answer. For example, a prompt template that can work well for all training examples is given the maximum generated probability in \cite{gao2020making}. 
Furthermore, language models can also be utilized to evaluate the quality of prompts. 
In \cite{davison2019commonsense}, the prompt with the highest probability given by an LM is selected, which indicates closer to the general expression that appears in the training dataset.

\subsubsection{Others}
Other criteria can be used to select the top one or the top-k prompt. For example, Shin \textit{et al.} \cite{shin2020autoprompt} regards the words that are estimated to have the largest performance improvement as the most crucial elements. 

\subsection{Tuning $P_T$}
Due to the continuous nature of LLMs, searching over discrete space is sub-optimal \cite{liu2021gpt}.
How can we further improve the performance once we obtain a prompt? 
Recent studies turn to optimizing the prompt as continuous embeddings. 


\begin{table*}[t!]
\caption{Summary of the prompt tuning methods}
\label{table:prompt_tuning}
\centering
\begin{tabular}{c|c|c|c}
\hline
Work  & Position & Length & Initialization  \\ 
\hline
\hline
prefix tuning \cite{li2021prefix} &  
prefix, infix & 200 (summarization), 10 (table-to-text) & random, real words  \\
prompt tuning \cite{lester2021power} & 
prefix & 1, 5, 20, 100, 150 & random, sampled vocab, class label     \\
p-tuning \cite{liu2021gpt} &  
hybrid & 3 (prefix), 3 (infix) & LSTM-trained     \\
DART \cite{zhang2021differentiable} &  
infix & 3 &  unused token in vocabulary   \\
OPTIPrompt \cite{zhong2021factual} &  
infix & 5, 10 & manual prompt     \\
dynamic \cite{yang2023dynamic} &  
hybrid, dynamic & dynamic & sampled vocab   \\
\hline
\end{tabular}
\label{table:prompt_tuning}
\end{table*}

The main idea is to learn a few continuous parameters, referred to as soft prompts, and these continuous parameters can be optionally initialized by the previously obtained discrete prompt. Li \textit{et al.} \cite{li2021prefix} first introduces a continuous task-specific `prefix-tuning' for generative tasks. 
Studies like \cite{lester2021power} and \cite{liu2021gpt} adopt a similar strategy and prove its effectiveness in various natural language understanding tasks. 
Following the above-mentioned studies, many improvements have been conducted to find better prompts, such as better optimizing strategies \cite{zhong2021factual}, better vector initialization \cite{gu2021ppt,hou2022metaprompting}, indicative anchors \cite{liu2021gpt} etc. Furthermore, studies like \cite{yang2023dynamic,li2021prefix,lester2021power} further point out that prompt position, length, and initialization all affect the performance of continuous prompts \cite{yang2023dynamic,li2021prefix,lester2021power} (Table \ref{table:prompt_tuning}). 
In this section, we summarize these factors as follows:

\begin{itemize}
    \item \textit{Different Position.} There are three different positions for autoregressive LM that the prompt can be inserted into, that is, the prefix $[PREFIX;X_T;Y]$, the infix $[X_T;INEFIX;Y]$, and the hybrid one $[PREFIX;X_T;INFIX;Y]$. 
    There is no significant performance difference between those positions. 
    \cite{li2021prefix} shows that prefix prompt sightly outperforms infix prompt, and the hybrid one is much more flexible than the others.

    \item \textit{Different Length.} There is no optimal length for all tasks, but there is always a threshold. 
    The performance will increase before reaching the threshold, then it will either plateau or slightly decrease. 
    
    \item \textit{Different Initialization.} A proper initialization is essential for the performance of the prompts and the performance of random initialization is usually unsatisfactory. Typical methods include initialized by sampling real word \cite{li2021prefix, lester2021power}, using class label \cite{lester2021power}, using discrete prompt \cite{zhong2021factual}, and using pre-trained based vector \cite{gu2021ppt,hou2022metaprompting}. 
    Furthermore, the manually designed prompts can provide a good starting point for the following search process.
\end{itemize}

\subsection{Trends for Prompt Template Engineering}
There are two trends in prompt template engineering:
\begin{itemize}
    \item Tend to have less human involvement, using automated methods rather than designing manually when constructing prompts.
    \item Tend to develop optimization-based techniques. The gradient-based searching method shows better performance than the derivative-free one in hard prompt construction while the soft prompt is more promising than the hard one. 
\end{itemize}
From the communication theory perspective, the development history of prompting template engineering reflects the trends of utilizing prompts with stronger expressive ability to better capture the user's intent. 

\section{Prompt Answering Engineering}
\label{sec:answer}

As illustrated in Fig. \ref{fig:parameters}(b), prompt answer engineering (PAE) aims to align LLMs outputs with the intended purpose. The use of PAE is motivated by the need to mitigate the gap between the capabilities of pre-trained LLMs and a large variety of requirements of different downstream tasks (see more discussion in Sect. 2). Technology-wise, PAE involves a set of methods that control admissible answer space and optimization mechanisms of  LLMs' output (see overview in Table~\ref{tab:answer-engin}).

\begin{table*}[t!] 
  \caption{Summary for the prompt answer engineering methods} 
  \centering
  \begin{tabular}{c|c|c|c}
        \hline
      Answer Space Type & Answer Mapping Method& Work & Task Type \\ 
      \hline
        \hline
\multirow{2}{*}{Optimizing the mapping} & Discrete Answer Space & \cite{schick2020automatically,schick2021exploiting, shin2020autoprompt, gao2021making} &   Classification \& regression    \\ \cline{2-4}
 & Continuous Answer Space & \cite{hambardzumyan2021warp} &   Classification  \\ \hline
Broadening the output  & discrete Answer Space & \cite{jiang2020how} &   Generation  \\ \hline
Decomposing the output & discrete Answer Space & \cite{chen-etal-2022-adaprompt} &   Classification \\ \hline
Manually Mapping & Pre-defined answer & \cite{petroni2019language,yin2019benchmarking,cui2021template} &   Generation  \\ \hline
  \end{tabular}
\label{tab:answer-engin}
\end{table*}

\subsection{Search for an Answer Space}
\subsubsection{Pre-defined Answer Space} This involves a set of pre-defined answers for the question-answering task, e.g., pre-defined emotions (``happiness'', ``surprise'', ``shame'', ``anger'', etc.) for the sentiment classification task. The model can then be trained to select the best answer from this pre-defined space. As an illustration, the answer space $P_A$ can be defined as the set of all tokens \cite{petroni2019language}, fixed-length spans \cite{jiang2020x}, or token sequences \cite{radford2019language}. Furthermore, in certain tasks like text classification, question answering, or entity recognition, answers are crafted manually as word lists that pertain to relevant topics \cite{yin2019benchmarking,cui2021template}.

\subsubsection{Discrete Answer Space}\label{sec:dis} 
Discrete answer space refers to a set of specific and distinct answer options that a language model can choose from when generating a response to a given prompt.

Specifically, the possible answers are limited to a fixed set of choices, such as a small number of named entities or keyphrases (e.g., the total choice of planet in the solar system is eight).
The model can then be trained to identify whether the correct answer is among this set of possibilities \cite{jiang2020how,schick2021exploiting,shin2020autoprompt}.

\subsubsection{Continuous Answer Space}\label{sec:conti} Continuous answer space refers to a scenario where the possible answers or responses are not restricted to a predefined set of discrete options. Instead, the answers can take on a range of continuous values or be any text, number, or value within a broader, unbounded spectrum \cite{hambardzumyan2021warp,nickel2018learning}. 

The model can then be trained to predict a point in this continuous space that corresponds to the correct answer.
\subsubsection{Hybrid Approach} This involves combining multiple methods to design the answer space, such as using a pre-defined list of entities for certain types of questions, but allowing for free-form text answers for other types of questions \cite{hou2020few}.
\begin{rmk}
Answer shapes summarized as follows are also needed in prompt answer engineering. In practice, the choice of answer shape depends on the desired outcome of the task.
\begin{itemize}
    \item Tokens: individual tokens within the vocabulary of a pre-trained Language Model (LLM), or a subset of the vocabulary..
    \item Span: short sequences of multiple tokens, often comprising a phrase or segment of text.
    \item Sentence: A longer segment of text that can encompass one or more complete sentences. 
\end{itemize}
\end{rmk}
\subsection{Search for an Answer Mapping}
There are several strategies to search for an answer mapping.
\subsubsection{Manually Mapping}
In many cases, the mapping from potential answers space $P_A$ to output $Y$ is obvious such that this mapping can be done manually. For instance, the answer is output itself for the translation task \cite{petroni2019language} such that the mapping is identity mapping; In addition, Yin et al \cite{yin2019benchmarking} designed 
related topics (``health'', ``food'', ``finance'', ``sports'', etc.), situations (``shelter'', ``water'', ``medical assistance'', etc.), or other possible labels. Cui et al. \cite{cui2021template} manually proposed some entity tags such as  "organization", "person" and "location", etc. for the Named Entity Recognition problem. 
\subsubsection{Broadening the answer $P_A$}
Broadening $P_A$ ($P'_A = B(P_A)$) is expanding the answer space to obtain a more accurate mapping. Jiang et al. \cite{jiang2020how} proposed a method to paraphrase the answer space $P_A$ by transferring the original prompt into other similar expressions. 
In their approach, they employed a back-translation technique by first translating prompts into another language and then translating them back, resulting in a set of diverse paraphrased answers. The probability of the final output can be expressed as $P(Y|x) = \sum_{y\in B(P_A)}P(y|x)$, where $B(Y)$ represents the set of possible paraphrased answers.
 
\subsubsection{Decomposing the output}
Decomposing $Y$ (D$(Y)$) aims to expand the information of $Y$, which makes it easier to look for a mapping $g_\theta$. For example, Chen et al. \cite{chen-etal-2022-adaprompt} decomposed the labels into several words and regarded them as the answer. Concretely, they decomposed label/output ``per:city\_of\_death'' into three separated words \{person, city, death\}. The probability of final output can be written as $P(y|x) = \sum_{y\in D(Y)}P(y|x)$.

\subsubsection{Optimizing the mapping.}
There exist two approaches to optimize the mapping function. The first approach is to generate the pruned space $\tilde{P_A}$ and search for a set of answers within this pruned space. Schick et al. \cite{schick2020automatically,schick2021exploiting} introduced a technique for generating a mapping from each label to a singular token that represents its semantic meaning. This mapping, referred to as a \textit{verbalizer} $v$, is designed to identify sets of answers. Their approach involves estimating a verbalizer $v$ by maximizing the likelihood w.r.t. the training data conditioned on the verbalizer $v$. 
Shin et al. \cite{shin2020autoprompt} proposed an alternative approach for selecting the answer tokens. They employed logistic classifiers to identify the top-k tokens that yield the highest probability score, which together form the selected answer. In addition, Gao et al. \cite{gao2021making} constructed a pruned set $\tilde{P_A}^c$ containing the top-k vocabulary words based on their conditional likelihood for each class $c$. As for the second approach, it investigates the potential of utilizing soft answer tokens that can be optimized through gradient descent. Hambardzumyan et al. \cite{hambardzumyan2021warp} allocated a virtual token to represent each class label and optimized the token embedding for each class along with the prompt token embedding using gradient descent.

\subsection{Trends for Prompt Answer Engineering}
There are two trends in prompt answer engineering:
\begin{itemize}
    \item Developing more robust and generalizable question-answering models that can handle more complex tasks and a broader range of inputs. For example, the answer space is some discrete spans at the beginning (see Sect. \ref{sec:dis}) and developed to the complex continuous space (see Sect. \ref{sec:conti}).
    \item There is also a focus on improving the quality and relevance of prompts to improve model performance. Specifically, several techniques have been explored, such as paraphrasing and pruning, after the direct mapping approach. More recently, optimization methods using gradient descent have been proposed to enhance accuracy.
\end{itemize}

The prompt answering engineering also shows a trend of exploring prompts to decode the machine language with less information loss, i.e., has a better understanding of the machine.

\section{Multiple Prompting Methods}
\label{sec:multi}
\begin{figure*}[t!]
    \centering
    \includegraphics[width=0.90\textwidth]{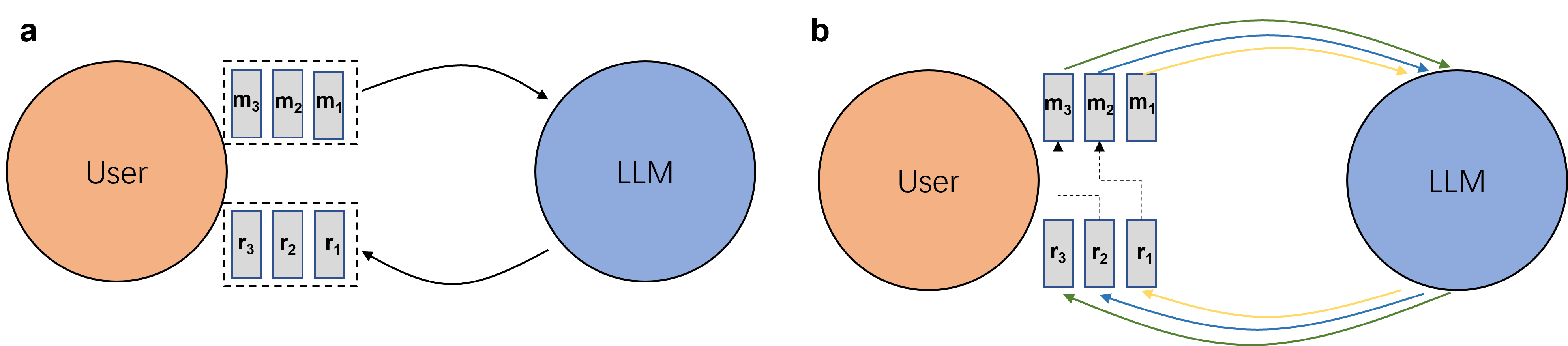}
    \caption{Overview of multiple prompting methods. (a) Multi-prompt methods utilize several similar prompts to produce a more stable result. (b) Multi-turn prompt methods produce the final result by aggregating responses from a sequence of prompts.}
    \label{fig:multi-prompt}
\end{figure*}

Multiple prompts can be utilized to further reduce the information mismatch during the encoding and decoding process. These methods can be categorized into two main types, namely ``multi-prompt engineering'' and ``multi-turn prompt engineering'', depending on the interrelationship of prompts (see Fig.~\ref{fig:multi-prompt}). Multi-prompt engineering is akin to an ensemble system, whereby each response serves as a valid answer, and responses from multiple prompts are aggregated to produce a more stable outcome. This type of method can be thought to extend the use of prompts in the spatial domain. On the other hand, multi-turn PE entails a sequence of prompts, whereby subsequent prompts depend on the response generated from previous prompts or the obtaining of the final answer relies on multiple responses. Consequently, this type of method can be viewed as an extension in the temporal domain.

\begin{table*}[t!]
\small
   \caption{Summary of the PE methods involving multiple prompts. NLU: Natural Language Understanding, NLG: Natural Language Generation.} 
   \centering
   \begin{tabular}{c|c|c|c|c}
        \hline
      & Method & NLU & NLG & Reasoning  \\ 
       \hline
        \hline
        \multirow{2}{*}{Multi-prompt} & Expanding $P_T$ & \cite{jiang2020can, lester2021power, hambardzumyan2021warp, qin2021learning} &   \cite{yuan2021bartscore} & - \\
        & Diversifying $P_A$ & - & - & \cite{wang2023selfconsistency,lewkowycz2022solving,wang2022rationaleaugmented,li2022advance,fu2023complexitybased}  \\
        & Optimizing $\theta$ & \cite{schick2020exploiting,schick2021it} & \cite{schick2021fewshot, gao2020making} & - \\
       \hline
        \multirow{2}{*}{Multi-turn prompt}  & Decomposing $P_T$ & - & \cite{perez2020unsupervised,min2019multihop,khot2021text} & \cite{zhou2022leasttomost,dua2022successive,creswell2022selectioninference,arora2022ask,khot2023decomposed,ye2023large,wu2022ai} \\
        & Refining $P_T$ & - & \cite{li2022selfprompting,ye2023explanation} & \cite{shum2023automatic,kojima2023large,ye2023explanation,wang2023selfconsistency,diao2023active,zhang2022automatic}   \\ 
        & Augmenting $P_T$ & - & \cite{yang2022re3,yang2022doc} & \cite{schick2023toolformer,shen2023hugginggpt,wang2022iteratively}   \\ 
        & Optimizing $\theta$ & - & \cite{perez2020unsupervised,min2019multihop} & \cite{wang2022iteratively,nye2021showa,zelikman2022star,taylor2022galactica}  \\ 
      \hline
   \end{tabular}
\label{table:multi-prompts}
\end{table*}

\subsection{Multi-prompt Engineering Methods}
Multi-prompt methods employ multiple prompts with similar patterns during the inference aiming to enhance information preservation. This method is closely associated with ensembling techniques~\cite{tingstacked, zhou2002ensembling,duh2011generalized}. Although the primary motivation is to exploit the complementary advantages of different prompts and reduce the expenses associated with PE, it can also be integrated with prompt-engineering techniques to further improve efficacy. From a communication theory perspective, multi-prompt engineering can be considered as sending multiple copies of the message to ensure the authentic delivery of data.

\subsubsection{Expanding $P_T$}

Expanding $P_T$ aims to cover a larger semantic space around the sender's true intention, and a more stable approximation of the target output, $\bar{X_A}$, can be obtained by aggregating the responses. 

Jiang \textit{et al.}~\cite{jiang2020can, lester2021power, hambardzumyan2021warp} propose to combine outputs of different prompts to get the final result for classification tasks. Qin \textit{et al.}~\cite{qin2021learning} incorporates multi-prompt ideas with soft ppromptsand optimizes weights of each prompt together with prompt parameters. Yuan \textit{et al.}~\cite{yuan2021bartscore} propose to use text generation probability as the score for text generation evaluation, and aggregate multiple results of different prompts as the final score.

\subsubsection{Diversifying $P_A$}

Different from expanding $P_T$ whose main goal is to leverage the input space around $P_T$, diversifying $P_A$ aims to exploit the various "thinking paths" of the LLM through sampling its decoder. This is especially effective for handling complex tasks, such as mathematical and reasoning problems.

Wang \textit{et al.}~\cite{wang2023selfconsistency} propose a self-consistency method based on the Chain-of-thoughts (CoT) which samples multiple reasoning paths and selects the most consistent answer by majority voting or weighted averaging. Lewkowycz ~\cite{lewkowycz2022solving} applied a similar idea to quantitative problems by combining multiple prompts and output sampling. Wang \textit{et al.}~\cite{wang2022rationaleaugmented} investigated various ensemble variants in reasoning problems and found that rational sampling in the output space is more efficient. These methods solely used the final answer as the selection criterion and did not exploit the generated rationals from various sampling paths. To take advantage of these intermediate results, Li \textit{et al.}~\cite{li2022advance} proposed to generate more diversified reasoning paths with multiple prompts and used a model-based verifier to select and rank these reasoning paths. Fu \textit{et al.}~\cite{fu2023complexitybased} introduced a complexity-based metric to evaluate reasoning paths and prioritize those with higher complexity in the aggregation. Weng \textit{et al.}~\cite{weng2023large} employed LLM to self-verify various reasonings by comparing predicted conditions using the generated reasonings to original conditions. The consistency score is then used to select the final result. Yao \textit{et al.}~\cite{yao2023tree} proposed the "Tree of Thoughts" to explore the intermediate steps across various reasoning paths, and used the LLM to evaluate the quality of each possible path.

\subsubsection{Optimizing $\theta$}
This line of work treats multiple prompts as a label generator to address the sample deficiency problem. Schick \textit{et al.}~\cite{schick2020exploiting} first proposes pattern-exploiting training (PET) that employs a knowledge distillation strategy to aggregate results from multiple prompt-verbalizer combinations (PVP). They first utilize PVP pairs to train separate models that generate pseudo-labels for unlabeled datasets. This extended dataset is then used to train the final classification model. Schick \textit{et al.}~\cite{schick2021few} extends this idea to the text generation task by using the generation probability of decoded text as the score. \cite{gao2020making} uses a similar method for automatic template generation. Schick \textit{et al.}~\cite{schick2021it} further expands PET with multiple verbalizers. This is achieved by introducing sample-dependent output space.

\begin{figure*}[t!]
    \centering
    \includegraphics[width=0.85\textwidth]{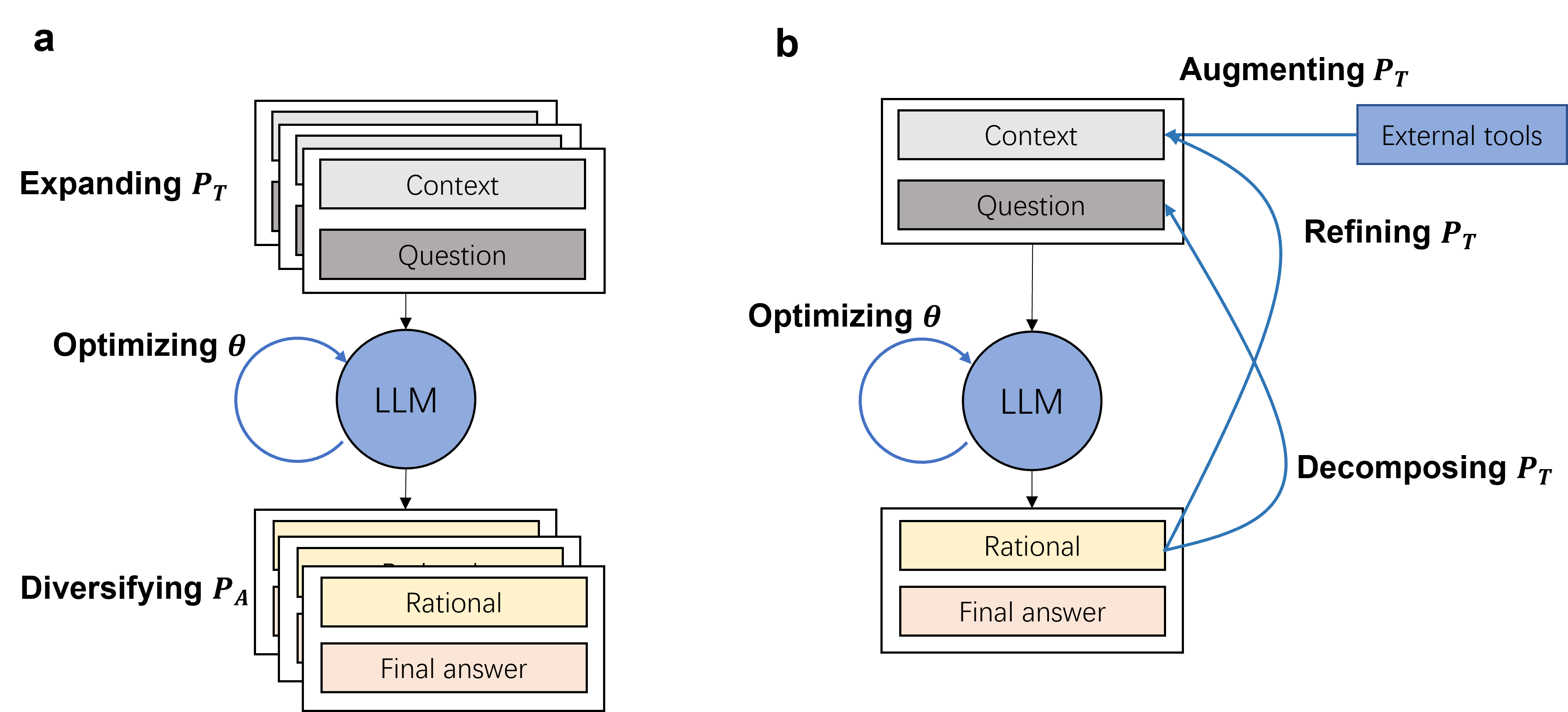}
    \caption{Schematic illustrations of multi-prompting methods. (a) Multi-prompt methods mainly employ ensemble-based methods. (b) Multi-turn prompt methods mainly leverage LLMs or external tools to provide clearer and more helpful context.}
    \label{fig:multi-prompt-methods}
\end{figure*}

\subsection{Multi-turn Prompt Engineering Methods}
Multi-turn prompt engineering methods involve decomposing the full prompting task into several sub-tasks, each addressed by a corresponding prompt. This process typically entails a sequence of encoding and decoding operations, where subsequent prompts may depend on the decoded message from previous prompts or each prompt is responsible for a sub-task. The outcome can be obtained either from the result of the last prompt or by aggregating the responses generated by all prompts. This strategy is designed to tackle challenging tasks, such as complex mathematical questions or reasoning tasks. It mainly involves two components: 1) decomposing $P_T$ into sub-tasks to reduce the difficulty of each sub-task; and 2) modifying $P_T$ to generate better intermediate results for later steps. These two components can help to bridge the gap between complex $X$ and $Y$.

\subsubsection{Decomposing $P_T$}
Decomposing $P_T$ is the first step in handling complex tasks, and a proper decomposition requires a good understanding of both the target task and the user's intention.
Yang \textit{et al.}~\cite{yang2022seqzero} decomposed SQL operations using fine-tuned few-shot models and untrained zero-shot models combined with predefined rules. However, ruled-based decomposition heavily relies on human experiences, so it is desirable to automate this step with LLMs. Min \textit{et al.}~\cite{min2019multihop} proposed an unsupervised method that utilizes a similarity-based pseudo-decomposition set as a target to train a seq2seq model as a question generator. The decomposed simple question is then answered by an off-the-shelf single-hop QA model. Perez \textit{et al.}~\cite{perez2020unsupervised} treats the decomposition in multi-hop reading comprehension (RC) task as a span prediction problem which only needs a few hundreds of samples. For each task, various decomposition paths are generated, with each sub-question answered by a single-hop RC model. Finally, a scorer model is used to select the top-scoring answer based on the solving path. Khot \textit{et al.}~\cite{khot2021text} proposed a text modular network leveraging existing models to build a next-question generator. The training samples are obtained from sub-task models conditioned on distant supervision hints.

With the emergent general ability of LLMs, instead of training a task-specific decomposition model, LLMs are used to fulfill decomposition tasks. Zhou \textit{et al.}~\cite{zhou2022leasttomost} proposed the least-to-most prompting method where hard tasks are first reduced to less difficult sub-tasks by LLM. Then answers from previous sub-problems are combined with the original task to facilitate subsequent question solving. Dua \textit{et al.}~\cite{dua2022successive} employs a similar idea and appends both question and answer from the previous stage to the subsequent prompt. Creswell \textit{et al.}~\cite{creswell2022selectioninference} proposed a selection-inference framework. It uses LLM to alternatively execute selecting relevant information from a given context and inferring new facts based on the selected information. Arora \textit{et al.}~\cite{arora2022ask} proposed to format the intermediate steps as open-ended question-answering tasks using LLMs. It further generates a set of prompt chains and uses weak supervision to aggregate the results. Khot \textit{et al.}~\cite{khot2023decomposed} proposed a modular approach for task decomposition with LLMs by using specialized decomposition prompts. Drozdov \textit{et al.}~\cite{drozdov2022compositional} introduced a dynamic least-to-most prompting method for semantic parsing tasks by utilizing multiple prompts to build a more flexible tree-based decomposition. Ye \textit{et al.}~\cite{ye2023large} uses LLMs as the decomposer for table-based reasoning tasks. LLMs are used for both sub-table extraction and question decomposition. Press \textit{et al.}~\cite{press2022measuring} proposed Self-Ask which decomposes the original task by repeatedly asking LLM if follow-up questions are needed. Wu \textit{et al.}~\cite{wu2022ai} proposed to build an interactive chaining framework with several primitive operations of LLM to provide better transparency and controllability of using LLMs.

\subsubsection{Refining $P_T$}

Refining $P_T$ aims to construct a better representation of $P_T$ based on the feedback from previous prompting results. This is especially important for multi-step reasoning, where the quality of generated intermediate reasonings has a critical impact on the final answer.

Following the success of the few-shot chain-of-thoughts (CoT) prompting method,
Kojima \textit{et al.}~\cite{kojima2023large} proposed a zero-shot CoT method that utilizes the fixed prompt 'Let's think step by step' to generate reasonings. These intermediate results are then fused with the original question to get the final answer. To select more effective exemplars, various methods are proposed. Li \textit{et al.}~\cite{li2022selfprompting} uses LLMs to first generate a pseudo-QA pool, then a clustering method combined with similarity to the question is adopted to dynamically select QA pairs from the generated QA pool as demonstration exemplars. Shum \textit{et al.}~\cite{shum2023automatic} leveraged a high-quality exemplar pool to obtain an exemplar distribution using a variance-reduced policy gradient estimator. Ye \textit{et al.}~\cite{ye2023explanation} employs self-consistency method\cite{wang2023selfconsistency} to generate pseudo-labels of an unlabeled dataset. The accuracy of these silver labels serves as the selection criterion of exemplars. To further reduce the search complexity of various combinations, additional surrogate metrics are introduced to estimate the accuracy. Diao \textit{et al.}~\cite{diao2023active} addresses this problem by using hard questions with human annotations as exemplars. The hardness is measured by the disagreement of results obtained by multiple sampling of the LLM. Zhang \textit{et al.}~\cite{zhang2022automatic} proposed automatic CoT methods. They introduced question clustering and demonstration sampling steps to automatically select the best demonstrations for the CoT template.

\subsubsection{Augmenting $P_T$}
Different from refining $P_T$ which mainly focuses on finding prompts that generate better intermediate results, augmenting $P_T$ leverages the exploitation of external information, knowledge, tools, etc. in the prompting. We present some examples in this field below, for more details we refer the reader to the specific survey \cite{mialon2023augmented}. 
Yang \textit{et al.}~\cite{yang2022re3} proposed a recursive reprompting and revision (3R) framework for long story generation leveraging pre-defined outlines. In each step, the context of the current status and the outline of the story is provided to the prompt to ensure better content coherence. Yang \textit{et al.}~\cite{yang2022doc} proposed to use more detailed outlines so that the story generation LLM can focus more on linguistic aspects. Information retrieved from other sources is also often used to augment $P_T$. Yao \textit{et al.}~\cite{yao2023react} gives the LLM access to information from Wikipedia. Thoppilan \textit{et al.}~\cite{thoppilan2022lamda} taught the LLM to use search engines for knowledge retrieval. More broadly, Paranjape \textit{et al.}~\cite{paranjape2023art} introduces a task library to enable the LLM using external tools. Schick \textit{et al.}~\cite{schick2023toolformer} trained the LLM to use various external tools via API. Shen \textit{et al.}~\cite{shen2023hugginggpt} utilized LLM as a central controller to coordinate other models to solve tasks.

\subsubsection{Optimizing $\theta$}
General LMs are not optimized for producing intermediate rationals or decomposing a complex task or question. Before the era of LLMs, these tasks require specifically trained LMs. Min \textit{et al.}~\cite{min2019multihop,perez2020unsupervised} trained an LM model for decomposing the original task into sub-tasks. Nye \textit{et al.}~\cite{nye2021showa} trains the LLM to produce intermediate steps stored in a scratch pad for later usage. Zelikman \textit{et al.}~\cite{zelikman2022star} utilized the intermediate outputs that lead to the correct answer as the target to fine-tune the LLM. Wang \textit{et al.}~\cite{wang2022iteratively} proposed an iterative prompting framework using a context-aware prompter. The prompter consists of a set of soft prompts that are prepared for the encoder and decoder of the LLMs respectively. Taylor \textit{et al.}~\cite{taylor2022galactica} employed step-by-step solutions of scientific papers in the training corpus, which enables the LM to output reasoning steps if required. 

\subsection{Trends for Multiple Prompting Methods}
Ensemble-based methods are easy to implement and flexible to incorporate with various strategies, e.g. expanding input space and aggregating output space. However, this brings limited advantages for complex problems whose final answers are hard to obtain directly, but rely heavily on the intermediate thinking steps. Therefore, multi-turn PE methods emerged. It essentially adjusts its input dynamically during the interaction based on the knowledge and feedback from the LLM or external tools. In this way, LLM can leverage more context and understand better the true intention of the user. Initially, specialized LLM are trained to handle planning and solving specific subtasks, this not only introduces extra training effort but also constrains the generalization capability of LLM. With the increasing understanding ability and larger input length of LLM, in-context learning becomes the preferred paradigm, which utilizes embedded knowledge and the capability of LLM to handle various tasks via prompting. This paradigm soon dominated because of its efficiency and flexibility.

There are two trends in multiple prompting engineering:
\begin{itemize}
    \item Developing an enhanced adaptive prompting strategy for LLM-based task decomposition is imperative. The extensive range and intricacy of tasks render human-based or rule-based task decomposition infeasible. While some studies have explored the use of LLM prompting to generate intermediate questions or actions for specific tasks, a comprehensive strategy is currently lacking.
    \item Enabling LLM to leverage tools without the need for fine-tuning is a crucial objective. By incorporating external tools, LLMs can address their limitations in specialized domains or capabilities. Previous studies have employed fine-tuning-based approaches to train LLMs in utilizing web search or other tools accessible through APIs.
\end{itemize}

From the communication theory perspective, multiple prompting methods evolved from the extension in the spatial domain (ensemble-based methods) into the temporal domain (mulit-turn), to better align the user's intention and LLM's capability by decomposing the user's request and leveraging external tools.

\section{Discussion}
\label{sec:dis}

Researchers have proposed several surveys to recapitulate the rapid advancements in the field of PE methods \cite{liu2023pre, qiao2022reasoning, lialin2023scaling, zhao2023survey, dong2022survey, lou2023prompt}. To name a few, Liu \emph{et. al} proposed a comprehensive survey about existing PE methods, which covers common aspects like template engineering, answering engineering, training strategies, applications, and challenges \cite{liu2023pre}. They reveal the development history of prompting learning and describe a set of mathematical notations that could summarize most of the existing studies. Furthermore, they consider prompt-based learning as a new paradigm that revolves around the way we look at NLP. In another survey \cite{qiao2022reasoning} that mainly focuses on the reasoning abilities (e.g., arithmetic, commonsense, symbolic reasoning, logical, multi-modal) of LLMs, Qiao \emph{et. al} summarized the studies that harness these reasoning abilities via advanced PE methods like chain-of-though and generated knowledge prompts. Additionally, some focused surveys cover specific topics like parameter-efficient fine-tuning (PEFT) LLMs using PE methods \cite{lialin2023scaling}. Different from the above-mentioned studies, we try to interpret existing PE methods from a communication theory perspective. 

Following this line of research, we also would like to discuss some potential challenges and future directions for PE methods, which could be divided into four categories including \emph{Reducing the Encoding Error}, \emph{Reducing the Decoding Error}, and \emph{Interactive and Multi-turn Prompting}.

\subsection{Reducing the Encoding Error}
\begin{itemize}
     
    \item \textit{Better Ranking Criteria}.
    One of the points of discrete prompts is that it's difficult to design and choose an optimal prompt, causing its instability. Although soft prompts partly addressed this problem, the discrete prompt is still very important because it has good interpretability and has been proven to be able to help soft prompts search effectively.
    Looking through the existing methods, we can find that accuracy-based criteria are resource-consuming, while LM-based log probability is not sufficient to evaluate the prompt. So a well-designed ranking criterion combined with a mass of auto-based generated prompts may be a good direction for the future.
    \item \textit{Task-agnostic Prompt}. Even though the prompt has been proven effective in many tasks such as classification and text generation, most of the existing work has to design a specific prompt for a given task, which makes it complex and complicated \cite{zhong2021adapting}. So how to generate a task-agnostic prompt or transfer the prompt to other fields quickly may be a challenging problem. Discrete(meta-learning \cite{reynolds2021prompt}) and continuous (decomposition \cite{madden2023few}) prompts are applied to tackle this issue. However, they are not well-optimized and can serve unseen tasks.
    \item \textit{Interpretability Issue}. 
    Recent studies show that those methods learning optimal prompts in continuous space can achieve better performance than in discrete space \cite{liu2023pre}. However, the generated `soft' prompts are difficult to read and understand, namely poor interpretability. Therefore, designing and improving soft prompts can be tough. Existing work \cite{lester2021power} tries to use the nearest words in embedding space to probe the 
    effect. However, the reasons excavated are not obvious. It remains to explore why this kind of prompt can work well and what causes the performance differences between different prompts. 
    \end{itemize}

\subsection{Reducing the Decoding Error}
\begin{itemize}
\item Privacy-preserving Methods \cite{abadi2016deep,gentry2009fully,yang2019federated}. To address privacy concerns on output, future research could focus on developing methods that preserve the privacy of the data used for training and inference. This could include techniques such as differential privacy, homomorphic encryption, and federated learning.

\item Human-in-the-loop Methods \cite{brennan1995interaction}. To improve the accuracy and relevance of prompt answer engineering methods, future research could focus on developing methods that incorporate human feedback and interaction. This could enable users to provide feedback and corrections to the generated answers and to refine the model over time.
\end{itemize}

\subsection{Interactive and Multi-turn Prompting}
    \begin{itemize}
    \item \textit{Transparency and Explainability}. Despite the recent popularity of LLMs, the lack of explainability of the outputs and transparency of the working mechanism makes LLMs less attractive in complex tasks that require high stability. The success of the chain-of-thoughts methodology shows the "thinking path" of LLMs can be evoked with proper indication. This property can be exploited to generate step-by-step task-solving procedures like scratch paper in exams so that the final answer can be better justified. \cite{wu2022ai} builds an interactive framework based on this idea and further involves human interaction for better controllability of the process. In addition to this online paradigm, LLMs can also be asked to explain the answer or decision afterward. \cite{wiegreffe2022reframingb} demonstrated that GPT-3 generates more favorable free-text explanations than crowdsourced.
    
    \item \textit{Interactive Multi-turn Prompt}. Though automation in prompting methods is highly wanted, humans in the loop can bring more controllability and supervision over the process, producing more reliable results. However, frequent human intervention will diminish the efficiency gained by using LLMs. Therefore, in addition to the granularity of decomposed tasks, it is also required to determine when to involve human feedback. This could be designed manually for each task, but it would be much more efficient if LLMs could plan these stages by themselves. 
    \end{itemize}

\section{Conclusion}
\label{sec:con}

This paper tries to provide an overview of existing prompting methods from a communication theory perspective. Towards this objective, we consider LLMs as a unified interface to achieve various NLP tasks and examine these prompt-based studies to reduce the information misunderstanding that appears in the different stages between users and LLMs during their interactions. We hope this survey will inspire researchers with a new understanding of the related issues in prompting methods, therefore stimulating progress in this promising area. 

\bibliographystyle{JCST}
\bibliography{bib}

\label{last-page}
\end{multicols}
\label{last-page}
\end{document}